\documentclass[10pt,twocolumn,letterpaper]{article}

\usepackage{iccv}              

\definecolor{iccvblue}{rgb}{0.21,0.49,0.74}
\usepackage[pagebackref,breaklinks,colorlinks,allcolors=iccvblue]{hyperref}
\usepackage{hyperref}

\def\paperID{10195} 
\def\confName{ICCV}
\def\confYear{2025}

\title{LiON-LoRA: Rethinking LoRA Fusion to Unify Controllable Spatial and Temporal Generation for Video Diffusion}

\author{Yisu Zhang$^{1,2}$\footnotemark[1]~ \footnotemark[3]~~, Chenjie Cao$^{2,3}$\footnotemark[1], Chaohui Yu$^{2,3}$, Jianke Zhu$^{1}$\footnotemark[2]\\
$^{1}$Zhejiang University
~~$^2$DAMO Academy, Alibaba Group ~~$^3$Hupan Lab \\
\small \texttt{\{zhyisu,jkzhu\}@zju.edu.cn}, ~~\texttt{\{caochenjie.ccj,huakun.ych\}@alibaba-inc.com},\\
}

\begin{document}
\maketitle

\renewcommand{\thefootnote}{\fnsymbol{footnote}}
\footnotetext[1]{Equal Contribution.  \footnotemark[2]Corresponding author.
\footnotemark[3]Work done during internship at DAMO Academy, Alibaba Group}

\begin{abstract}

Video Diffusion Models (VDMs) have demonstrated remarkable capabilities in synthesizing realistic videos by learning from large-scale data. 
Although vanilla Low-Rank Adaptation (LoRA) can learn specific spatial or temporal movement to driven VDMs with constrained data, achieving precise control over both camera trajectories and object motion remains challenging due to the unstable fusion and non-linear scalability. 
To address these issues, we propose \textbf{LiON-LoRA}, a novel framework that rethinks LoRA fusion through three core principles: \textbf{Linear scalability}, \textbf{Orthogonality}, and \textbf{Norm consistency}. 
First, we analyze the orthogonality of LoRA features in shallow VDM layers, enabling decoupled low-level controllability. 
Second, norm consistency is enforced across layers to stabilize fusion during complex camera motion combinations. 
Third, a controllable token is integrated into the diffusion transformer (DiT) to linearly adjust motion amplitudes for both cameras and objects with a modified self-attention mechanism to ensure decoupled control. 
Additionally, we extend LiON-LoRA to temporal generation by leveraging static-camera videos, unifying spatial and temporal controllability. 
Experiments demonstrate that LiON-LoRA outperforms state-of-the-art methods in trajectory control accuracy and motion strength adjustment, achieving superior generalization with minimal training data. Project Page: https://fuchengsu.github.io/lionlora.github.io/

\end{abstract}    
\begin{figure*}
    \begin{center}
    \includegraphics[width=1\linewidth]{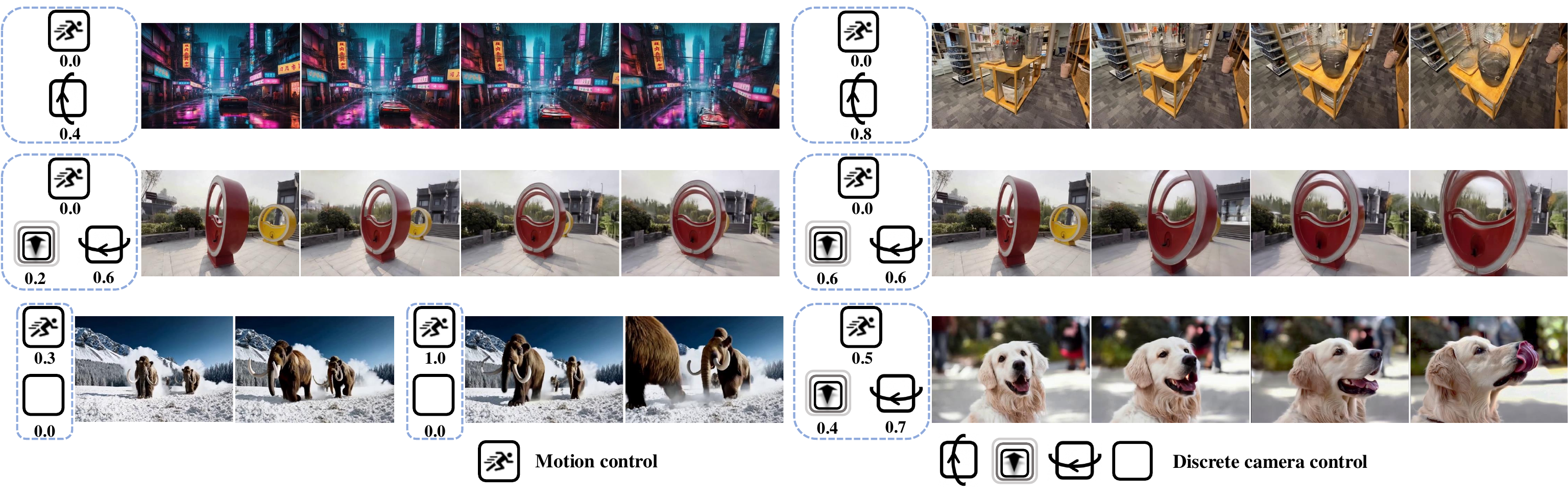}
    \end{center}
    \vspace{-0.15in}
    \caption{\textbf{Camera and motion control results of our LiON-LoRA.} LiON-LoRA can linearly control both camera trajectory and object motion in videos generated by the video diffusion model. Furthermore, based on LoRA fine-tuning, LiON-LoRA achieves excellent generalization with minimal training data.
    \label{fig:teaser}}
    \vspace{-0.15in}
\end{figure*}

\section{Introduction}

Rapid advancement of foundational video diffusion models (VDMs)~\cite{he2022latent, chen2023videocrafter1, wang2023modelscope, wang2023lavie, yang2024cogvideox, videoworldsimulators2024, xing2024dynamicrafter, Blattmann2023, chen2023seine, zhang2023i2vgen, zhang2024pia} has enabled them to learn realistic interactions from the physical world, resulting in impressive video outputs.
Using vast amounts of video data available on the Internet, VDMs have significant potential to understand 3D spatial and even 4D temporal information in the real world~\cite{hu2024animate, sun2024dimensionx, jiang2025animate3d, bahmani2024ac3d, yu2024viewcrafter, jiang2024consistentd, yin20234dgen}. Consequently, reactivating the potentials of VDMs with good controllability would largely promote many downstream applications, such as augmented reality experiences, interactive video editing, and simulation of virtual environments.
Despite the abundance of in-the-wild video data, the lack of proper annotations presents a challenge in taming VDMs for 3D and 4D generation, particularly in terms of unified control of both camera and motion strength, as illustrated in \Cref{fig:teaser}.

Recent works integrated explicit camera presentations~\cite{Wang2024Motionctrl, He2024Cameractrl, yang2024direct, xu2024camco, zheng2024cami2v} or motion guidance~\cite{jain2024peekaboo, wang2024videocomposer, Wang2024Motionctrl, yin2023dragnuwa} to control VDM. 
However, these manners usually undermine the generalization of VDMs, leading to inferior synthesis quality as a compromise.
Thus, equipping VDMs with effective controllability while working with limited labeled video data remains a significant challenge.
Fortunately, Parameter-Efficient Fine-Tuning (PEFT)~\cite{ding2023parameter} alleviates this challenge by reducing the dependency on extensive annotated training data while employing lightweight model weights, thus mitigating the degradation of generalization in VDMs.
Low-Rank Adaptation (LoRA)~\cite{hu2021lora} has garnered considerable attention among these PEFT approaches due to its flexibility and versatility. LoRA utilizes low-rank factorized matrices with a minimal number of trainable parameters, making it widely applicable in text-to-image and image-to-image personalization~\cite{zhong2024multi, shah2023ZipLoRA, ouyang2025k}.
Additionally, AnimateDiff~\cite{hu2024animate} employs 8 distinct LoRAs to drive the respective camera controls within the VDM, while DimensionX~\cite{sun2024dimensionx} further decouples the temporal and spatial factors through unified LoRAs for advanced 4D generation. 

Despite the pioneering advances mentioned above, implicitly driving video generation via LoRA fails to achieve as precise control as explicit injection, which could be attributed to two factors: 1) \emph{Unstable LoRA fusion.} Unlike high-level LoRA fusion techniques used for tasks such as style and subject editing~\cite{shah2023ZipLoRA, yang2024loracomposer, huang2023lorahub, ouyang2025k}, we find that combining different camera control LoRAs suffers from obvious instability, such as abrupt changes in direction, as illustrated in \Cref{fig:param}.
2) \emph{Non-linear LoRA scalability.} While the conventional approach of adjusting the multiplying coefficients of LoRA features is adequate for modifying high-level properties, such as style transfer, it fails to precisely control camera movements, which demands a deeper low-level spatial understanding.

The underlying issue of these challenges lies in the tendency of most previous works to focus on high-level LoRA fusion, neglecting an in-depth discussion about inherent LoRA mechanisms to quantify low-level controllability.
In this paper, we rethink LoRA mechanisms from three aspects, including \textbf{Linear scalability}, \textbf{Orthogonality}, and \textbf{Norm consistency}, and further present \textbf{LiON-LoRA}, a framework with unified spatial and temporal controls for VDM, fully compatible with all aforementioned features.

Formally, we first analyze the orthogonality of distinct camera LoRAs and find that LoRA features from shallow layers of VDM exhibit low feature correlation, which is theoretically sufficient to decouple low-level controllability.
Moreover, we recognized that distinct camera LoRAs possess disparate magnitudes (norms), leading to unstable LoRA fusions. Therefore, we normalize the output norms from each layer of the LiON-LoRA, ensuring norm consistency and prominently enhancing the stability of LoRA fusion for complex camera motion combinations.
Subsequently, we achieve the linear scalability of LiON-LoRA by inserting a dedicated \textit{scaling token} within the LoRA training. Specifically, we concatenated an additional global token to the diffusion transformer (DiT) feature sequence, linearly encoding the motion amplitudes of both cameras and objects. 
When fusing multiple LoRAs, we adjust the receptive field of self-attention to attend to separate scaling tokens from different LoRAs, enjoying decoupled linear control.
Additionally, we collect extensive video clips with static camera positions, extending the linear scalability of LiON-LoRA into temporal control. This unifies both 3D and 4D generation within VDM.
Experiments of camera control with both simple and complicated trajectories demonstrate that LiON-LoRA outperforms other methods, offering superior controllability and generalization with minimal training samples. We also evaluate the Pearson correlation, denoting that LiON-LoRA can effectively control motion strength for 4D generation.

We summarize the key contributions of LiON-LoRA as
\begin{itemize}
    \item Unlike previous LoRA works that focus primarily on high-level fusion, we revisit the LoRA fusion in low-level camera control and present LiON-LoRA, acquiring linear scalability, orthogonality, and norm consistency. 
    \item We propose a simple yet effective way to normalize the LiON-LoRA feature's norm to enable stable camera movement fusion. 
    \item We introduce a scaling token for LiON-LoRA, which enables linear scalability to adjust the influence of camera and object movements.
    \item Although LiON-LoRA is highly training-efficient, it successfully unifies both spatial and temporal controllability, facilitating impressive 4D generation.
\end{itemize}

\section{Related Work}
\label{sec:2_related_work}

\noindent\textbf{Image-to-Video Generation.} Recent years have witnessed remarkable advances in video generation technology, significantly propelling the development of content synthesis fields. Early research efforts~\cite{he2022latent,wang2023modelscope,wang2023lavie,girdhar2023emu,Blattmann2023,wang2023modelscope,chen2023videocrafter1,hu2024animate,chen2024videocrafter2} 
Pioneered text-to-video (T2V) generation by adapting pre-trained text-to-image (T2I) models through architectural adjustments or fine-tuning strategies. Subsequently, commercial video generation engines \eg, Sora~\cite{videoworldsimulators2024}, Gen-3~\cite{runwaygen3}, and Kling~\cite{kling} have demonstrated expanded capabilities in T2V synthesis, along with extended functionalities that include image-to-video (I2V) conversion and specialized video effect generation. 
These closed-source systems typically employ comprehensive video generation pipelines incorporating extensive pre-processing and post-processing modules.
Recent open source initiatives have improved methodological transparency and accessibility for research communities~\cite{xing2024dynamicrafter,zheng2024open,lin2024open,yang2024cogvideox,kong2024hunyuanvideo}, providing modular frameworks for video synthesis. 
While both CogVideoX~\cite{yang2024cogvideox} and HunyuanVideo~\cite{kong2024hunyuanvideo} adopt the MMDiT structure, a modified Transformer variant employing full attention mechanisms~\cite{esser2024scaling}. In contrast, Open-Sora leverages the DiT backbone~\cite{Peebles2023,Yu2024,Ma2024}, demonstrating competitive performance through spatial-temporal modeling in latent space.

\noindent\textbf{Camera Controllable Video Generation.} 
Controlled generation has gained increasing attention in both image~\cite{Zhang2023, Mou2024, peng2024controlnext, ye2023ip, wu2024spherediffusion, song2024moma, wu2024ifadapter} and video generation~\cite{guo2023sparsectrl, peng2024controlnext, chen2024motion, hu2024animate}, with control signals evolving from simple text descriptions to include audio~\cite{tang2023anytoany,tian2024emo,he2024co}, identity~\cite{chefer2024still,wang2024customvideo,wu2024customcrafter}, user interaction patterns~\cite{yin2023dragnuwa}, and camera trajectory specifications~\cite{chen2024motion, wu2024motionbooth, yang2024direct, li2024generative, namekata2024sg}. Among these, text-based camera control is one of the simplest yet most effective approaches, leveraging natural language descriptions to guide camera motion. Early research~\cite{Guo2023, Blattmann2023} used fine-tuning techniques, such as LoRA, to parse control signals from text inputs. Furthermore, training-free methods~\cite{hu2024motionmaster, jain2024peekaboo} have been proposed to generate coarse-grained camera movements, though with limited precision. Although these text-based approaches enable camera trajectory control, they often struggle with precise control and exhibit limited controllability.
To address the challenge of camera control, some methods~\cite{Wang2024Motionctrl,He2024Cameractrl,li2025realcam, yang2024direct} explicitly inject camera pose as an additional input into the model, while others~\cite{zheng2024cami2v, xu2024camco} leverage epipolar attention to improve the multiview consistency. 
Furthermore, some works~\cite{yu2024viewcrafter,feng2024i2vcontrol} take advantage of recent advances in monocular depth estimation models~\cite{depth_anything_v1, depth_anything_v2} to reconstruct 3D point clouds from images as explicit control, allowing accurate camera movement for free-view rendering. 
However, these methods require intensive training preparation that involves the extraction of motion trajectories from large-scale video data. 
Additionally, such explicit controls struggle to handle dynamic scenes due to their overly rigid conditions.
Despite these explorations, the challenges of achieving precise and controllable camera and object motion using easily constructed, small-scale datasets persist. Our method aims to address these limitations.

\noindent\textbf{Motion Control in Video Generation}
Object motion control has emerged as a critical interaction channel for video manipulation, with recent methods predominantly employing bounding-box annotations or drag operations as control signals. Bounding box-based approaches~\cite{jain2024peekaboo, wang2024videocomposer, zheng2024cami2v} enable object trajectory specification through sequential bounding boxes that outline target positions across frames. Alternative drag-based methods~\cite{yin2023dragnuwa, wu2024draganything, Wang2024Motionctrl} leverage sparse point trajectories to manipulate object movements. These methods suffer from limited trajectory capacity and complex user interactions, while still generating implausible motions from inadequate real-world dynamics modeling.
In contrast to these trajectory-based control mechanisms, our method introduces a parameterized motion scale that jointly encodes motion properties in 3D space. Besides, this control paradigm avoids complex interactions while preserving the plausibility of natural motions. 

\section{Method}
\label{sec:3_method}

\noindent\textbf{Overview.}
We first briefly review LoRA and its usage for camera control in \Cref{sec:periminary}.
Based on this formulation, we then analyze the challenges of vanilla LoRA fusion for camera control (\Cref{sec:rethink}). 
Subsequently, we propose LiON-LoRA with orthogonality and norm consistency (\Cref{sec:orth_norm}), as well as linear scalability (\Cref{sec:linear_scale}) respectively. Finally, we introduce the training-free fusion of LiON-LoRA in \Cref{sec:lora_fusion}.

\subsection{Preliminary: Discrete LoRA Camera Control}
\label{sec:periminary}

Complex camera movements in 3D space can be decomposed into combinations of fundamental motion primitives. Through LoRA~\cite{hu2021lora} fine-tuning, we demonstrate that VDMs can effectively adapt to basic motion patterns using only a minimal set of reference videos with limited training iterations, as evidenced in \Cref{tab:data_abla}. 
Therefore, we modify open-source CogVideoX ~\cite{yang2024cogvideox} by injecting LoRA modules into the linear layers of the transformer blocks. This lightweight adaptation enables precise control over camera motion patterns while preserving the model's original generative capabilities. The parameter update process for each motion primitive is formalized as:

\begin{equation}
\Delta W_i = A_iB_i,  \quad A_i \in \mathbb{R}^{d\times r}, B_i \in \mathbb{R}^{r\times k},
\end{equation}

\begin{equation}
W_i={W}^{\text{base}}_i + \lambda \Delta W_i,
\end{equation}

where $i$ indexes distinct motion primitives, $r$, $d$, and $k$ represent the LoRA rank, input, and output dimensions, respectively. In addition, ${W}^{\text{base}}_i$ and $\Delta W_i$ represent the original CogVideoX weights and low-rank updates introduced by LoRA, while $\lambda$ controls the intensity of LoRA injection to achieve the final model weights $W_i$.

In this paper, three types of fundamental camera primitives are included:
(1) \emph{horizontal/vertical offset}, 
(2) \emph{forward/backward movement}, 
(3) \emph{orbital rotation around the scene center}. 
Each primitive requires dedicated LoRA weight training, which requires motion-specific video data preparation. 
To achieve videos to learn distinct camera control LoRAs, we render fixed and stable trajectories through 3D Gaussian Splatting (3DGS)~\cite{kerbl20233dgs} trained on the multi-view dataset, DL3DV~\cite{ling2024dl3dv}. We adaptively decide the moving distance according to the medium depth of each scene. More details are discussed in the supplementary.

\subsection{Rethinking LoRA Fusion for Camera Control}
\label{sec:rethink}

\begin{figure}
    \begin{center}
    \includegraphics[width=1\linewidth]{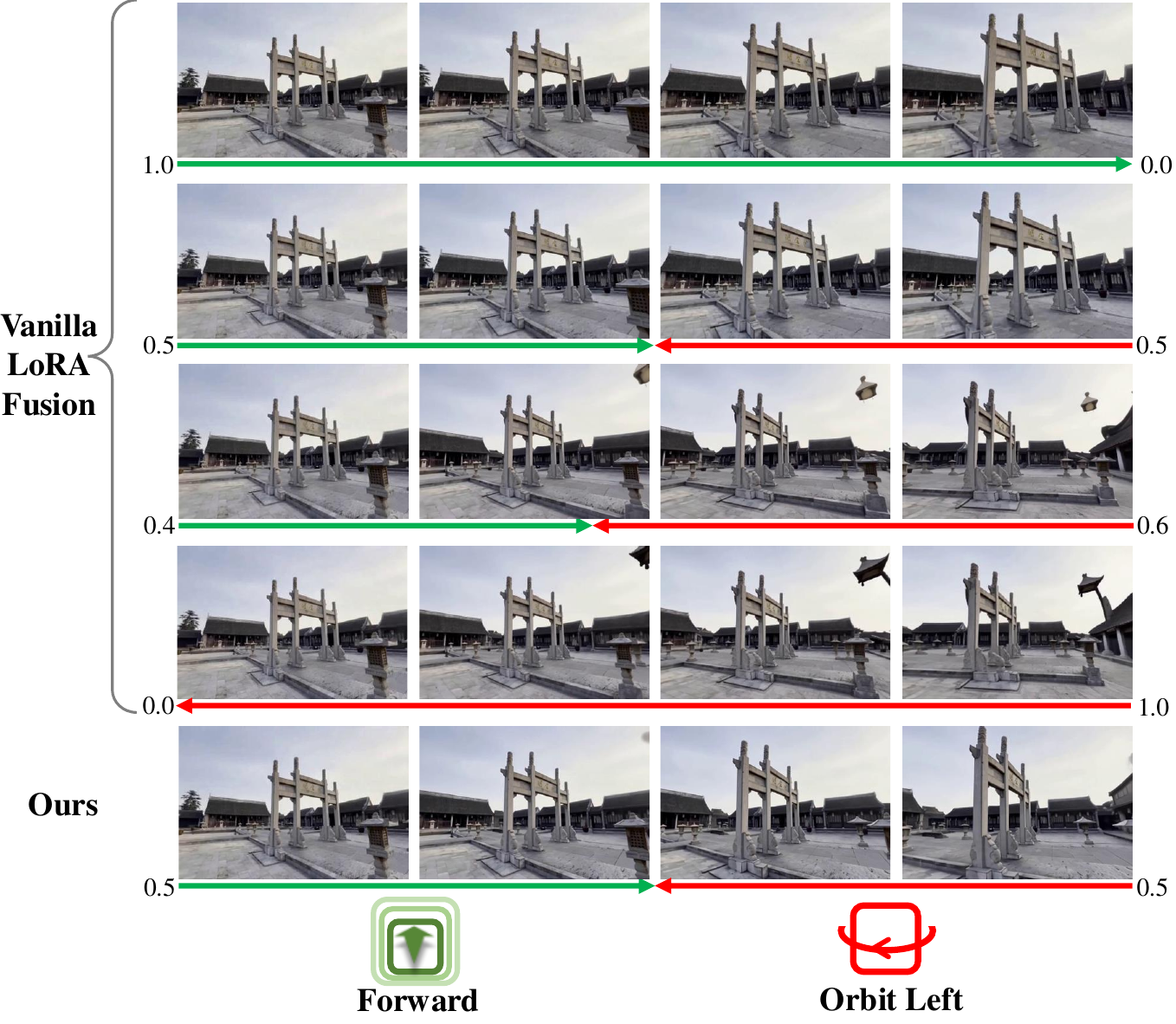}
    \end{center}
    \vspace{-0.25in}
    \caption{\textbf{Comparison between the vanilla LoRA fusion and ours.}
    The vanilla fusion of ``forward'' and ``orbit left'' LoRA controls with adjusted adapter scales suffers from abrupt changes.
    While our LiON-LoRA enhanced with norm consistency enjoys a much smoother combination without parameter adjusting.\label{fig:param}}
    \vspace{-0.15in}
\end{figure}

Pre-trained foundational generative models typically enable linear combinations of multiple LoRA weights to achieve multitask capabilities, where the fusion strategy used between different LoRA modules is particularly significant. The most straightforward method for LoRA fusion involves linearly combining multiple LoRAs $\Delta{W}_i$ through coefficients $\lambda_i, i\in\{1,2,...,k\}$ as: 
\begin{equation}
\Delta{W} = \sum_{i=1}^k \lambda_i\Delta{W}_i,
\end{equation}
where $\Delta{W}$ denotes the weight of the fused LoRA module.
Although directly linearly combining LoRA weights performs well for high-level controllable tasks, such as style and subject fusion~\cite{zhong2024multi, shah2023ZipLoRA, ouyang2025k}, this vanilla strategy does not manage satisfactory low-level control, such as camera movements fusion, as illustrated in \Cref{fig:param}.
Specifically, the LoRA fusion of the vanilla camera suffers from abrupt changes (\eg, suddenly switching from forward movement to orbit left) rather than a smooth progression in blending these two camera motions, even when the combining coefficients are carefully adjusted.

\begin{figure}
    \begin{center}
    \includegraphics[width=1\linewidth]{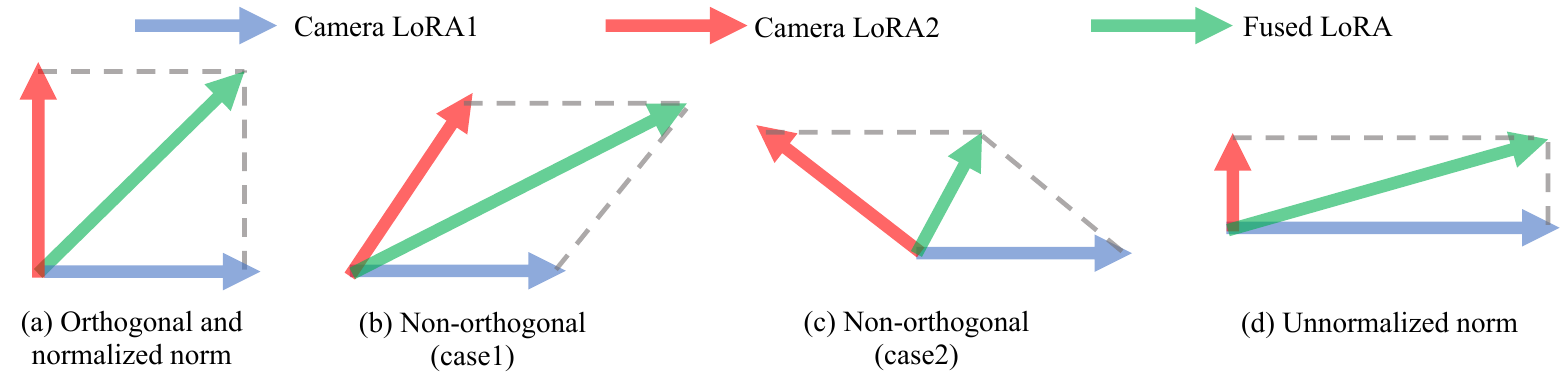}
    \end{center}
    \vspace{-0.15in}
    \caption{\textbf{Illustration of the dilemma for LoRA fusion in camera control.}
    We show the 2D toy case to demonstrate the importance of orthogonality and norm consistency.
    Non-orthogonality results in undesired enhancements or conflicts of fusions (b, c), while significantly different norms lead to unstable fusion (d).\label{fig:orth}}
    \vspace{-0.15in}
\end{figure}

To address these issues, we rethink the LoRA fusion from the perspective of vector space principles, including orthogonality, norm consistency, and precise linear scalability.  
\textbf{Linear scalability}: Although learning LoRA primitives is an implicit way to control camera trajectories, explicit linear scalability is essential to achieve accurate camera control in linear space, which cannot be achieved simply by adjusting the LoRA coefficient $\lambda$. This property not only facilitates flexible camera control but also enhances the LoRA fusion process by enabling smoother movements.
\textbf{Orthogonality}: When multiple tasks share a base model, high-level correlations among LoRA features may lead to unwanted enhancements or conflicts, as shown in \Cref{fig:orth}(b) and \Cref{fig:orth}(c), compromising the characteristics of each LoRA during fusion. Thus, ensuring the orthogonality of the camera LoRAs should serve as the foundation for an effective camera LoRA fusion.
\textbf{Norm consistency}: Since different camera control LoRAs are trained independently, their low-rank matrices exhibit divergent Frobenius norms. This disparity results in disproportionate amplification or suppression of specific LoRA weights during the combination, allowing one module to dominate over others, as shown in \Cref{fig:orth}(d). 
In this paper, we introduce \textbf{LiON-LoRA}, a method that encompasses all three properties, effectively decoupling camera control capabilities within the parameter space while maintaining their combined expressive power.

\subsection{Orthogonality and Norm Consistency}
\label{sec:orth_norm}

\begin{figure}
    \begin{center}
    \includegraphics[width=1.0\linewidth]{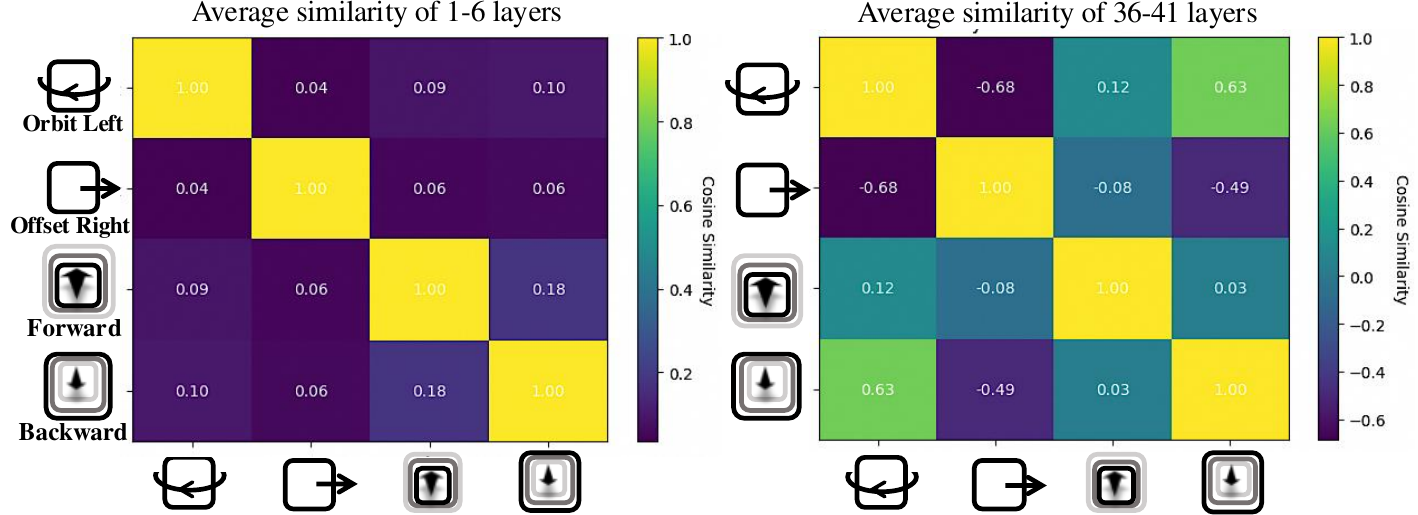}
    \end{center}
    \vspace{-.15in}
    \caption{\textbf{Cosine similarity of different camera LoRAs.} In the shallow layers, LoRAs maintain orthogonality, while in the later layers, coupling effects begin to emerge.\label{fig:cossim}}
    \vspace{-.05in}
\end{figure}

\noindent\textbf{Orthogonality.}
As mentioned above, the orthogonality should be the base to get the complicated trajectories combined from discrete ones.
As visualized in~\Cref{fig:cossim}, our layer-wise similarity analysis reveals some factors of camera LoRAs' orthogonality: LoRA outputs in earlier transformer blocks exhibit strong orthogonality, with an average cosine similarity of 0.06±0.06, while deeper blocks show a notable increase in correlation. 
This phenomenon suggests that camera LoRAs have achieved almost orthogonal characteristics in the shallower layers.
As demonstrated in~\cite{bahmani2024ac3d,sun2024dimensionx}, the low-frequency camera controls are primarily encoded within the initial transformer blocks of VDMs, with subsequent layers emphasizing high-frequency generation. This observation aligns with our theoretical analysis, \ie, the LoRA features corresponding to different camera trajectories exhibit good orthogonality in early blocks, suggesting that the orthogonality should be sufficient to handle the low-frequency camera control.

\begin{figure}
    \begin{center}
    \includegraphics[width=1.0\linewidth]{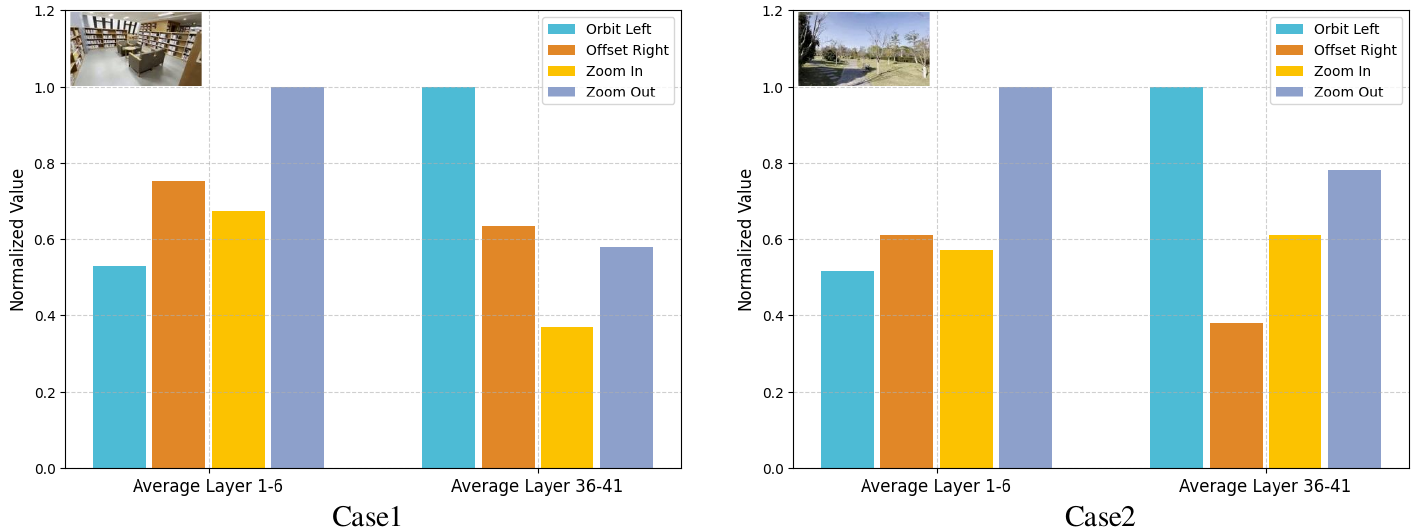}
    \end{center}
    \vspace{-.18in}
    \caption{\textbf{Norm Consistency of different LoRAs.} The norms of outputs from different camera control LoRAs vary significantly for different samples, which occur in both early and later layers, reflecting that LoRA weights fail to ensure norm consistency.\label{fig:norm}}
    \vspace{-.20in}
\end{figure}

\noindent\textbf{Norm Consistency.}
As illustrated in \Cref{fig:norm}, we observe significant norm disparities among different LoRAs related to various camera trajectories and transformer blocks across different samples.
These disparities stem from variations in the underlying distribution of the training data.
Notably, the norm discrepancies are not uniform across the transformer blocks, making it challenging to establish a pre-set adapter scale that effectively balances the LoRA outputs.
To address this issue, we propose a post-normalization approach tailored to individual LoRA outputs to ensure a balanced contribution across different LoRAs as follows:
\begin{equation}
\Delta \hat{W}_i = \frac{\alpha}{\left\|\Delta W_i\right\|} \Delta W_i,
\end{equation}
where $\alpha$ serves as a unified scaling factor for $k$ different LoRAs, typically set to $\alpha=\sum(\left\|\Delta W_i\right\|) / k$ for balanced norm preservation.  
This simple yet effective layer-wise norm normalization significantly improves the smoothness of LoRA fusion, as verified in \Cref{fig:param}.

\subsection{Scaling Controllable Video Diffusion}
\label{sec:linear_scale}

\textbf{Linear Scalability.}
Our empirical pilot studies reveal that adjusting the adapter value $\lambda$ of each camera LoRA during training fails to enable explicit linear control.
We summarize the limitations of conventional LoRA scaling as follows.
(1) $\lambda$ is not integrated as the inputs of VDMs, without explicitly affecting other video and textual features to control generations.
(2) Increasing the value of $\lambda$ undermines the overall generalization of the base VDM, contradicting the primary purpose of LoRA.
(3) $\lambda$ serves to determine the intensity of LoRA injection, facilitating high-level feature fusion. However, this adjustment struggles to achieve fine-grained precision in controlling camera trajectories.

To address these limitations, we adopt a flexible control mechanism by incorporating an additional \emph{scaling token} into the visual and textual token sequence to explicitly represent the scale of camera motion. 
Specifically, we define a scaling value $\mathcal{S}\in[s,1]$ to present the amplitude of each camera movement, where $s$ denotes the minimal scaling value. 
Then we render corresponding videos from DL3DV which contains 600 frames for each video clip.
Consequently, we could prepare training pairs of $\mathcal{S}$ and $V=49$ video frames uniformly sampled from the first $600\cdot \mathcal{S}$ frames of each clip, while the minimal $s=49/600$.
To improve the fine-grained representation of $\mathcal{S}$, we employ the Fourier-based positional embedding $\gamma(\cdot)$ before sending it to the linear projection as:
\begin{equation}
\gamma(S) = \sum_{j=0}^{J-1}[\sin(2^j\pi S), \cos(2^j\pi S)],
\end{equation}
\begin{equation}
\mathcal{E} = \textrm{linear}(\gamma(S)),
\end{equation}
where $j\in\{0,...,J-1\}$ denote the frequency indice, and $\mathcal{E}$ is the scaling token. 
Subsequently, $\mathcal{E}$ is further concatenated with the visual and textual token sequence ${H}$ as:
\begin{equation}
{H}' = [{H}; \mathcal{E}] \in \mathbb{R}^{(n+1)\times d},
\end{equation}
where $n,d$ denotes the original sequential length and channels for VDM's transformer blocks, respectively.
Note that we use specific project linear to encode $\mathcal{S}$ for different camera LoRAs, thereby avoiding coupling that could hinder orthogonality. LiON-LoRA remains parameter-efficient since the additional trainable parameters associated with the linear projection are negligible.
As shown in \Cref{fig:scaling}, the scaling token outperforms the adapter value. Our systematic comparison demonstrates that the scaling token achieves superior and more stable linear scalability.

\begin{figure}[h]
    \begin{center}
    \includegraphics[width=1\linewidth]{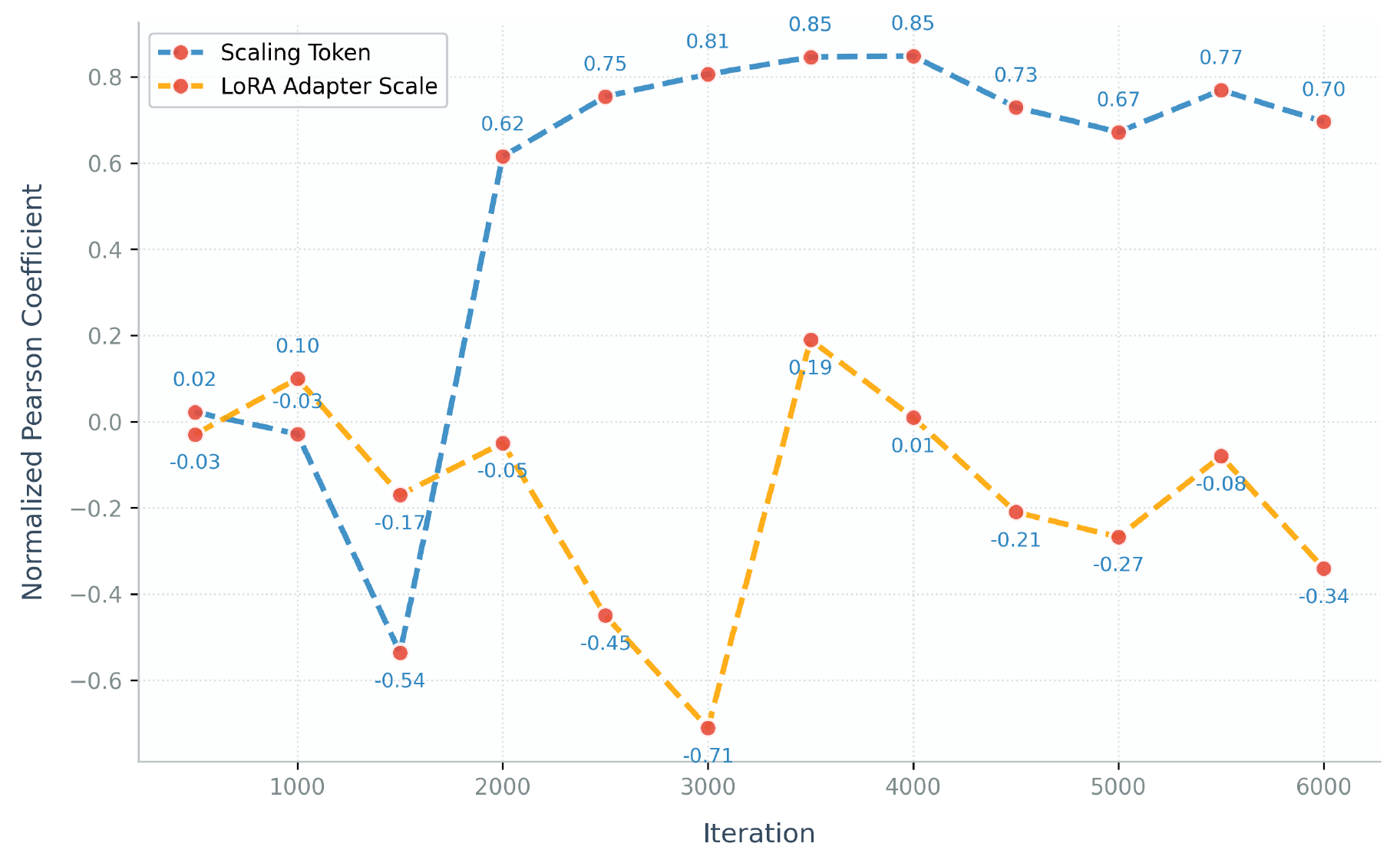}
    \end{center}
    \vspace{-.15in}
    \caption{\textbf{Pearson correlation of motion strength controllability.} We quantify motion controllability through Pearson correlation analysis between optical flow magnitudes and two control types (scaling scale vs LoRA adapter scale). While the adapter scale exhibits poor motion linearity, our scaling token enables precise control over object motion strength.\label{fig:pearson}}
    \vspace{-.15in}
\end{figure}

\begin{figure}[h]
    \begin{center}
    \includegraphics[width=1\linewidth]{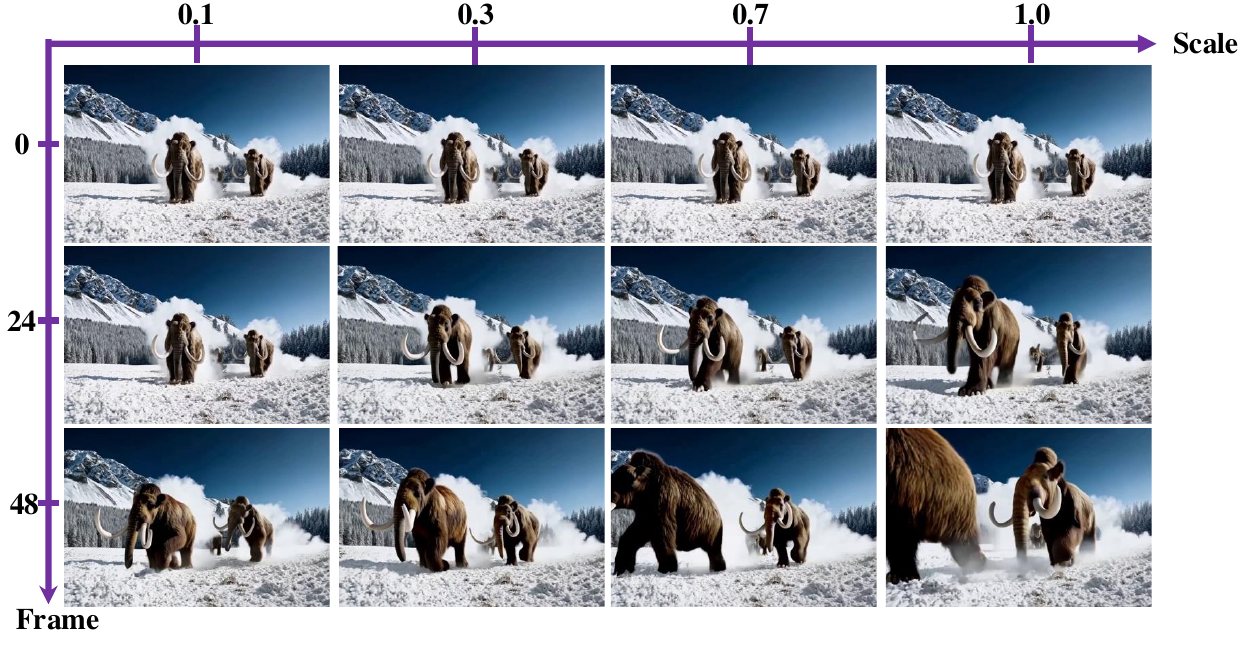}
    \end{center}
    \vspace{-.15in}
    \caption{\textbf{Illustration of motion strength scaling.} Our LiON-LoRA enables the control of object motion amplitude through the scaling token without complex bounding box annotations or drag-based interactions. Furthermore, this parameter-driven approach ensures that the generated videos exhibit both temporal continuity and visual consistency.\label{fig:motion}}
    \vspace{-.15in}
\end{figure}

\noindent\textbf{Motion Strength Scaling.}
The proposed LoRA scaling paradigm can seamlessly be extended to control the scale of motion strength.
Instead, the scaling value $\mathcal{S}$ is defined to present the speed of object motion. Similar to the linear scalability in controllable camera motion, scaling value $S \in [s,1]$ represents the magnitude of object motion. Different from the setting in camera motion, the maximum frame of the object motion video is 240, resulting in the minimal $s=49/240$. 
To ensure that the scaling token focuses exclusively on motion strength—unimpeded by undesirable camera movements—we train the motion strength scaling only on videos with static camera positions.
To validate the effectiveness of our proposed motion scaling, we compute the Pearson correlation between $\mathcal{S}$ and the optical flow magnitude over consecutive frames, as illustrated in \Cref{fig:pearson}. 
The results demonstrate that our method achieves rapid convergence during the early stages of fine-tuning LiON-LoRA. In contrast, the vanilla adapter scaling fails to establish a meaningful correlation for effective control.
Moreover, we show the qualitative results in \Cref{fig:motion}, the scaling token successfully controls the motion strength with linear scalability.

\subsection{Training Free Fusion}
\label{sec:lora_fusion}

\begin{figure}
    \begin{center}
    \includegraphics[width=1\linewidth]{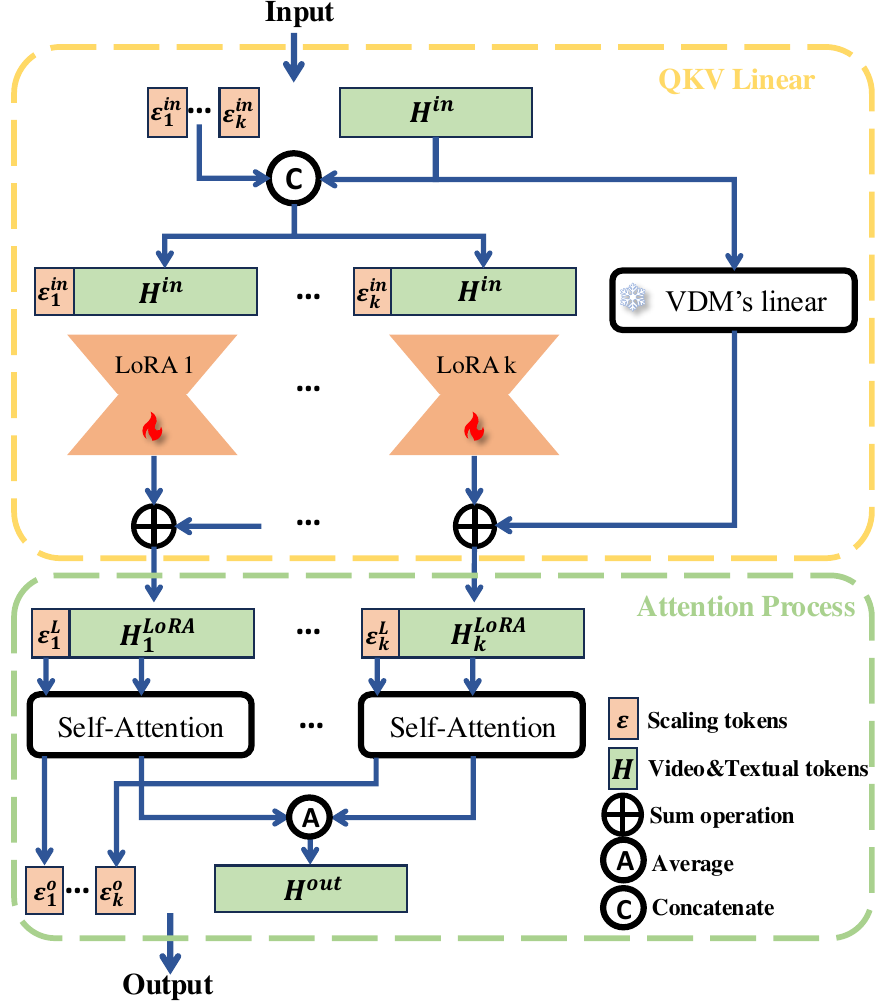}
    \end{center}
    \vspace{-0.2in}
    \caption{\textbf{Pipeline of multiple LiON-LoRA fusion.} 
    LiON-LoRA first splits scaling tokens and then assigns them to the individual LoRA encoding and attention process.
    Finally, we average $k$ groups of output visual and textual tokens as $H^{out}$, while $k$ output scaling tokens are spatially concatenated.
    \label{fig:block}}
    \vspace{-0.15in}
\end{figure}

The proposed LiON-LoRA achieves both orthogonality and norm consistency without altering the existing LoRA fusion process.
However, incorporating new scaling tokens during the LoRA fusion could potentially impact the video sequence and compromise the orthogonality of LoRA outputs, thereby inhibiting independent control.
Therefore, we propose a novel fusion framework that enables independent scaling control for multiple LoRAs as illustrated in \Cref{fig:block}.
Given $k$ scaling values $\{S_1,...S_k\}$, each value is processed through dedicated encoding layers to corresponding scaling tokens $\{ \mathcal{E}_1, ...,\mathcal{E}_k\}$. 
These scaling tokens are then concatenated with original hidden states of video sequence $H$ to form ${H}^{\prime}$:
\begin{equation}
{H}^{\prime}=[{H}; \mathcal{E}_1; \ldots; \mathcal{E}_k] \in \mathbb{R}^{(n+k) \times d}.
\end{equation}
To maintain individual control without feature coupling, each LoRA output is designed to attend exclusively to a specific scaling token $\mathcal{E}_i$ within its own independent attention subspace, denoted as ${H_i}^{\prime}=[{H}; \mathcal{E}_{i}]$.
After the self-attention, different scaling tokens are spatially concatenated while the hidden states are averaged. 
The resulting output tokens ${H}^{out} \in \mathbb{R}^{(n+k) \times d} $ then propagate through the subsequent blocks, ensuring that control is properly disentangled.
Notably, LiON-LoRA fusion is training-free and highly versatile, allowing for multi-trajectory camera fusions and hybrid camera-motion controls without any joint training.

\section{Experiments}
\label{sec:4_exp}

\subsection{Experiment Setup}

\noindent\textbf{Dataset.}
We select the large-scale, multi-category scene dataset DL3DV~\cite{ling2024dl3dv} as the training dataset of camera control. For each basic camera trajectory, we reconstruct the scene using 3DGS~\cite{kerbl20233dgs} and render videos by moving the virtual camera along pre-defined trajectory patterns, including horizontal/vertical offset, forward/backward movement, and orbital rotation around the scene center.
Ultimately, we reconstruct 100 scenes and render the corresponding videos for each fundamental camera primitive.
To train controllable object motion, we collected 172 videos from the Internet, which contain videos with flexible object movements while the camera remains static. 
These videos with static cameras are filtered by heuristic processing within boundary tracking points detected from Cotracker~\cite{karaev2024cotracker}.
We further employed the optical flow to filter videos with too small motions.
This design facilitates the decoupling of camera and object motion, enabling precise control over both modalities. 
For quantitative performance evaluation, we selected 100 samples from DL3DV for each motion primitive and conducted evaluations on both basic and complex camera poses.
For qualitative results, we conduct comprehensive experiments across diverse data sources including DL3DV, web images, and generated images.

\begin{figure*}[h]
    \begin{center}
    \includegraphics[width=1\linewidth]{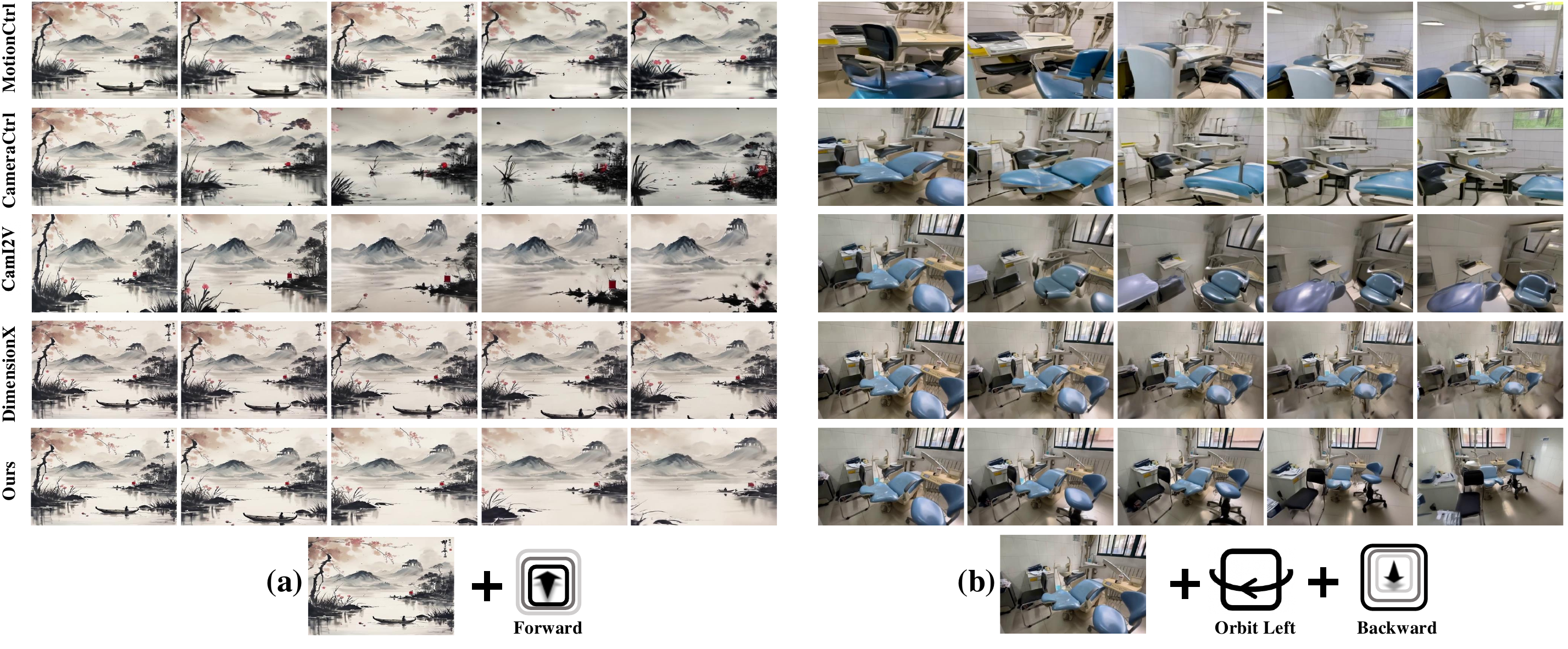}
    \end{center}
    \vspace{-.2in}
    \caption{\textbf{Qualitative comparisons on camera pose control.} 
    (a) Our model achieves diverse styles with strong generalization and adaptability. Existing methods (MotionCtrl~\cite{Wang2024Motionctrl}, CameraCtrl~\cite{He2024Cameractrl}, CamI2V~\cite{zheng2024cami2v}) struggle with inconsistent scene geometry and hallucinations across styles. 
    The re-implemented DimensionX~\cite{sun2024dimensionx} has limited camera movement amplitude and lacks linear control. 
    (b) Under complex trajectories, our method effectively integrates diverse camera movements while maintaining high quality. Others face limitations in large-scale motion and generalization, while DimensionX suffers from LoRA conflicts, losing some camera movements (orbit left).\label{fig:result}}
    \vspace{-.15in}
\end{figure*}

\noindent\textbf{Implementation Details.}
Our LiON-LoRA is based on CogVideoX~\cite{yang2024cogvideox}, a VDM trained on 49-frame sequences. This LoRA-based training approach can be seamlessly adapted to other video diffusion models, such as HunYuan~\cite{kong2024hunyuanvideo} and DynamicCrafter~\cite{xing2024dynamicrafter} with its flexibility. During training, we supervise the diffusion model using videos at a resolution of $480 \times 720$, following the CogVideoX pre-training scheme. We set the LoRA rank at 256 and fine-tuned the LoRA layers for 4,000 steps with a learning rate of $5 \times 10^{-4}$. We train LiON-LoRA and variants on 4 NVIDIA H20 GPUs with a batch size of 16.
For inference, we employ DDIM sampling with 50 steps, and the classifier-free guidance scale is set to 6. 

\noindent\textbf{Metrics and Baselines.}
Following CameraCtrl~\cite{He2024Cameractrl}, we evaluate the accuracy of camera pose predictions using three metrics. RotErr (rotation error), TransErr (translation error), and ATE (absolute trajectory error). To compute these metrics with the structure-from-motion method, following CamI2V~\cite{zheng2024cami2v} we use GLOMAP~\cite{pan2024glomap} to convert multiview images into camera poses for the predicted videos. Additionally, we assess video generation quality using FVD~\cite{Unterthiner2018}.
We compare our method with several state-of-the-art baselines, including DimensionX~\cite{sun2024dimensionx}, MotionCtrl~\cite{Wang2024Motionctrl}, CameraCtrl~\cite{He2024Cameractrl}, and Cami2V~\cite{zheng2024cami2v}. 

\begin{table}
  \renewcommand{\arraystretch}{1.2}
  \centering
  \caption{\textbf{Quantitative results of basic camera poses.} The DimensionX-S$^*$ indicates the re-implemented version of S-Director of DimensionX~\cite{sun2024dimensionx}. }
  \vspace{-0.3cm}
  \resizebox{\linewidth}{!}{
  \begin{tabular}{c|cccc}
    \toprule
    Method & \textbf{RotErr~$\downarrow$} & \textbf{TransErr~$\downarrow$}  & \textbf{ATE~$\downarrow$} & \textbf{FVD~$\downarrow$} \\
    \midrule
    CogVideoX~\cite{yang2024cogvideox}  &  4.974 & 0.765 & 0.980 & 387.6   \\
    MotionCtrl~\cite{Wang2024Motionctrl}      & 2.254 & 0.269 & 0.408 & 290.9  \\
    CameraCtrl~\cite{He2024Cameractrl}  & 1.737 & \underline{0.192} & 0.458 & 218.9  \\
    CamI2V~\cite{zheng2024cami2v}      & \underline{1.033}  & 0.215 & 0.370 & 294.6  \\
    DimensionX-S$^*$~\cite{sun2024dimensionx}  & 1.223 & 0.201 & \underline{0.359} & \underline{193.3} \\
    Ours        & \textbf{0.776} & \textbf{0.167} & \textbf{0.295} & \textbf{136.0}  \\

    \bottomrule
  \end{tabular}}
  \vspace{-12pt}
  \label{tab:base_motion}
\end{table}

\begin{table}
  \renewcommand{\arraystretch}{1.2}
  \centering
  \caption{\textbf{Quantitative results of complex fused camera poses.}}
  \vspace{-0.3cm}
  \resizebox{\linewidth}{!}{
  \begin{tabular}{c|cccc}
    \toprule
    Method & \textbf{RotErr~$\downarrow$} & \textbf{TransErr~$\downarrow$}  & \textbf{ATE~$\downarrow$} & \textbf{FVD~$\downarrow$} \\
    \midrule
    CogVideoX~\cite{yang2024cogvideox}      & 4.104 & 0.717 & 0.842 & 383.2  \\
    MotionCtrl~\cite{Wang2024Motionctrl}  & 2.063 & 0.317 & 0.426 & 357.2  \\
    CameraCtrl~\cite{He2024Cameractrl}      & 1.947 & \underline{0.201} & 0.440 & 261.8   \\
    CamI2V~\cite{zheng2024cami2v}  & \textbf{0.924} & 0.217 & \underline{0.398} & 285.6  \\
    DimensionX-S$^*$~\cite{sun2024dimensionx}  & 1.510 & 0.264 & 0.405 & \underline{240.7}  \\
    Ours        & \underline{1.044} & \textbf{0.197} & \textbf{0.345} & \textbf{172.8}  \\

    \bottomrule
  \end{tabular}}
  \vspace{-12pt}
  \label{tab:fuse_motion}
\end{table}

\begin{figure}[h]
    \begin{center}
    \includegraphics[width=1\linewidth]{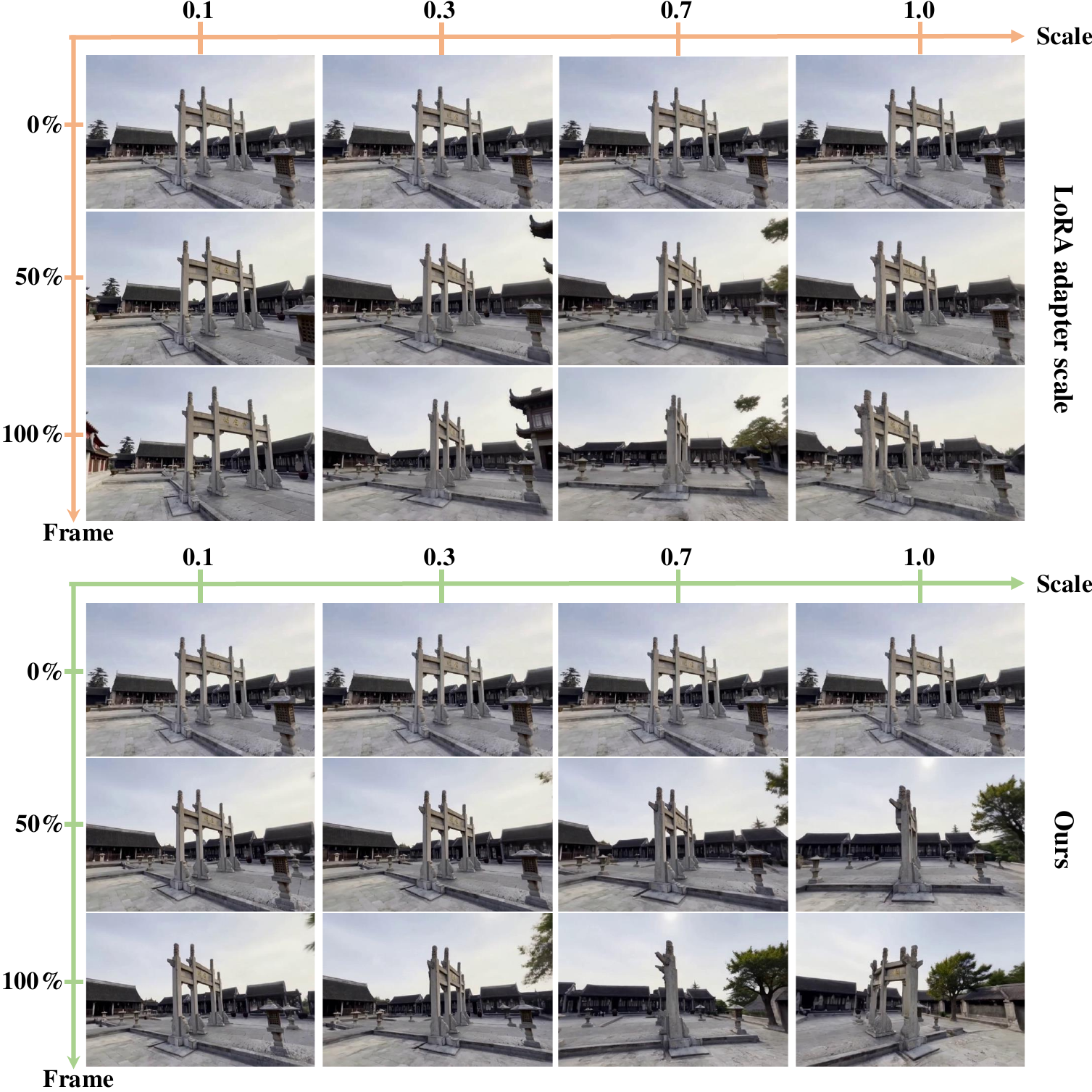}
    \end{center}
    \vspace{-.2in}
    \caption{\textbf{Linear scalability of LoRA.} Vanilla LoRA adjusts injection strength via the global ``adapter scale" $\lambda$, which lacks precision in camera motion control. Specifically, an insufficient adapter scale fails to control the camera of VDM, while a large adapter scale may lead to poor quality. In contrast, our LiON-LoRA enables linear controllability with better VDM generalization.
    }
    \vspace{-.15in}
    \label{fig:scaling}
\end{figure}

\subsection{Qualitative Results}
The qualitative results are presented in \Cref{fig:result}. It can be seen that our model outperforms MotionCtrl, CameraCtrl, and CamI2V in terms of higher resolution, better consistency, less hallucinated content, and longer sequence generation. Furthermore, our approach requires only iterations of $4,000$ fine-tuning, significantly fewer than the at least $50,000$ steps typically needed by other competitors. This efficient training strategy helps preserve the generalization of the VDM. 
A particularly notable comparison is with the DimensionX baseline, where our model achieves linear controllability and multi-motion fusion under the same foundation model, CogvideoX.
Furthermore, CameraCtrl, MotionCtrl, and CamI2V may suffer from hallucinated content and temporal flickering due to limitations in their base models and potential overfitting to specific training datasets.

\subsection{Quantitative Results}
The evaluation encompasses both fundamental camera poses and their complex combinations to assess motion control fidelity. As demonstrated in \Cref{tab:base_motion} and \Cref{tab:fuse_motion}, our method achieves significant improvements in both camera controllability and visual quality. 
For fundamental camera poses, quantitative results reveal superior performance across all metrics: Rotation Error (RotErr) $\downarrow 24.8\%$, Translation Error (TransErr) $\downarrow 13.0\%$, and Average Trajectory Error (ATE) $\downarrow 21.5\%$. 
The enhanced control capability stems from introducing the scaling token that enables precise control of camera trajectories. Currently, due to the efficient training architecture, the reduced number of fine-tuning iterations enables our framework to maintain enhanced temporal consistency during long-form video generation (49 frames vs. 16-frame), demonstrating a $37.8\%$ reduction in FVD compared to CamI2V.
As for complex camera motion combinations, our method also achieves superior performance except RotErr metric, as demonstrated in~\Cref{tab:fuse_motion}.

\subsection{Ablation Study}

\noindent\textbf{Dataset Size and Training Iterations.} 
Compared to existing approaches that process Plücker~\cite{Plucker1828} embeddings as global positional embedding to inject camera parameters, which typically require large-scale datasets and extensive fine-tuning (10k+ iterations), our method achieves precise linear controllability with significantly fewer training samples. We conducted comparative experiments on the ``orbit left" camera motion trajectory using 7k versus 100 training instances, as illustrated in ~\Cref{tab:data_abla}. The results demonstrate that our model attains satisfactory controllability after only 4k iterations with minimal data.

\begin{table}
  \renewcommand{\arraystretch}{1.2}
  \centering
  \caption{\textbf{Ablation study on dataset size and training iterations.}}
  \vspace{-0.3cm}
  \resizebox{\linewidth}{!}{
  \begin{tabular}{cc|cccc}
    \toprule
    Samples & Iteration & \textbf{RotErr~$\downarrow$} & \textbf{TransErr~$\downarrow$}  & \textbf{ATE~$\downarrow$} & \textbf{FVD~$\downarrow$} \\
    \midrule
    100 & 4k   & 0.802 & 0.143 & 0.331 & \textbf{175.3}   \\
    100 & 10k  & \textbf{0.710} & 0.146 & \textbf{0.327} & 184.6  \\
    7k & 4k    & 1.227 & 0.252 & 0.390 & 232.5  \\
    7k & 10k   & 0.896 & \textbf{0.140} & 0.341 & 202.8  \\

    \bottomrule
  \end{tabular}}
  \vspace{-12pt}
  \label{tab:data_abla}
\end{table}

\noindent\textbf{Scaling Token.}
Our method achieves precise linear controllability by injecting control signals through additional scaling tokens in the transformer architecture. As compared to the vanilla adapter scaling approach of LoRA in~\Cref{fig:scaling}, our framework not only enables accurate control but also effectively decouples the intensity of LoRA injection from control signals. The scaling token preserves the VDM's generalization capability while maintaining a more stable generation quality across different control magnitudes.
\section{Conclusion}
\label{sec:5_conclusion}

We present LiON-LoRA, a parameter-efficient framework designed for unified control over spatial and temporal generation in VDMs from three key principles: linear scalability, orthogonality, and norm consistency.
Particularly, we find that the camera control LoRAs exhibit strong orthogonality in the shallow layers of VDM, while we further normalize the norm of LoRA outputs for norm consistency, facilitating smooth camera LoRA fusion.
Moreover, we propose the scaling token to explicitly control both camera movements and motion strength with linear scalability. 
A tailored training-free LoRA fusion framework is proposed to achieve independent controls across various LoRA modules and scaling tokens.
Extensive experiments validate the effectiveness and efficiency of LiON-LoRA with minimal training data.
{
    \small
    \bibliographystyle{ieeenat_fullname}
    \bibliography{main}
}

\end{document}